\RequirePackage[T1]{fontenc}
\documentclass[letterpaper, 10 pt, conference]{ieeeconf}
\IEEEoverridecommandlockouts
\usepackage{graphicx}
\usepackage{float}
\usepackage{amsmath}
\usepackage{amssymb}
\usepackage{booktabs}
\usepackage{makecell}
\usepackage{xspace}
\usepackage{cite}
\usepackage{url}
\makeatletter
\let\NAT@parse\undefined
\makeatother
\usepackage[unicode,colorlinks=true,allcolors=blue]{hyperref}

\newcommand{\pp}{\textsc{PonderPounce}}
\newcommand{\ponder}{\textsc{Ponder}}
\newcommand{\pounce}{\textsc{Pounce}}
\newcommand{\pizeroonehalf}{\ensuremath{\pi_{0.5}}\xspace}
\newcommand{\taskins}{\mathrm{Ins}}
\newcommand{\demoreason}{\mathrm{DR}}
\newcommand{\subgoalreason}{S}
\newcommand{\tyes}{T_{\mathrm{Yes}}}
\newcommand{\tno}{T_{\mathrm{No}}}
\newcommand{\maintablestyle}{%
    \small
    \setlength{\tabcolsep}{4pt}%
    \renewcommand{\arraystretch}{1.08}%
}
\newsavebox{\tablecontentbox}
\newcommand{\fittable}[1]{%
    \sbox{\tablecontentbox}{#1}%
    \ifdim\wd\tablecontentbox>\linewidth
        \resizebox{\linewidth}{!}{\usebox{\tablecontentbox}}%
    \else
        \usebox{\tablecontentbox}%
    \fi
}
\newcommand{\myparagraph}[1]{\noindent\textbf{#1}}

\title{\LARGE \bf
\pp{}: A Pretrained MLLM as an\\
Episode Context Engine for Robot Control
}

\author{%
    \authorblockN{%
        Suhwan Choi$^{1,2,*}$ \quad
        Jaeyoon Jung$^{1,*}$ \quad
        Sungkyung Kim$^{3}$ \quad
        Yunsung Lee$^{1,\dagger}$ \quad
        Youngjae Yu$^{2,\dagger}$}
    \authorblockA{%
        $^{1}$MAUM.AI \qquad $^{2}$Seoul National University \qquad $^{3}$Georgia Institute of Technology}
    \thanks{$^{*}$Equal contribution. $^{\dagger}$Co-corresponding authors.}
}

\hypersetup{
    pdftitle={PonderPounce: A Pretrained MLLM as an Episode Context Engine for Robot Control},
    pdfauthor={Suhwan Choi, Jaeyoon Jung, Sungkyung Kim, Yunsung Lee, Youngjae Yu}
}

\makeatletter
\IEEEaftertitletext{%
\begin{minipage}{\textwidth}
    \centering
    \def\@captype{figure}%
    \includegraphics[width=0.99\textwidth]{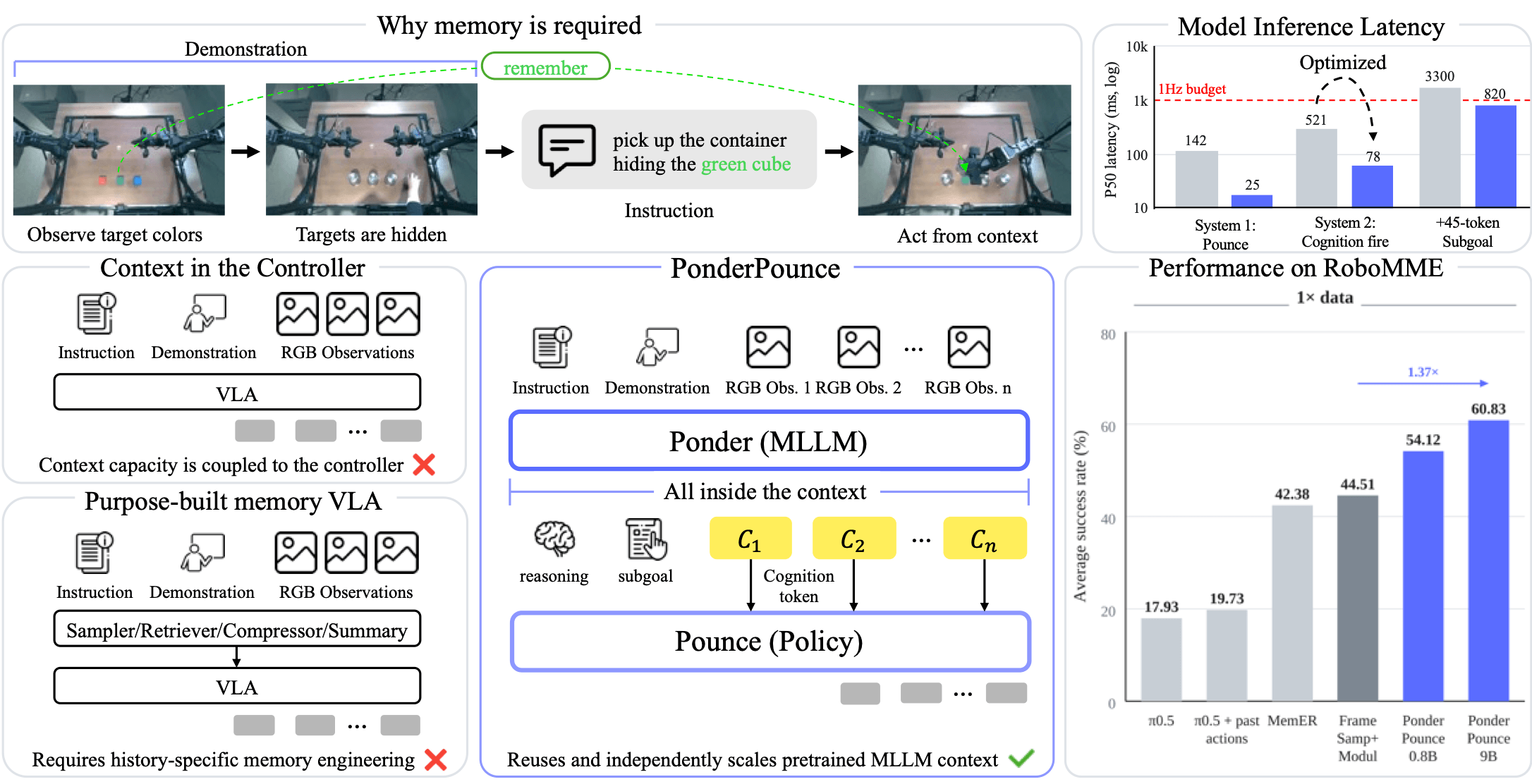}
    \caption{\textbf{Pretrained MLLM context as scalable robot memory.} The example shows our real-robot TrackCube task (Sec.~\ref{sec:exp_real}). Unlike designs processing context within the controller or through purpose-built memory, \pp{} retains history in \ponder{}'s context and asynchronously routes cognition to \pounce{}, keeping context processing off the action path and System~2 scalable without changing the controller architecture.}
    \label{fig:overview}
\end{minipage}%
\vspace{\textfloatsep}%
\vspace{-1.5em}%
}
\makeatother

\begin{document}
\maketitle
\thispagestyle{empty}
\pagestyle{empty}

\begin{abstract}
    Multimodal large language models (MLLMs) can integrate long visual histories and infer behavior from a few examples, yet vision-language-action models rarely use this capacity as episode memory. Instead of a purpose-built memory module, \pp{} reuses an MLLM's native causal context. \ponder{}, a pretrained System~2 MLLM, integrates episode history and demonstrations to produce continuous cognition. \pounce{}, a System~1 action model, asynchronously conditions control on the newest cognition and its age. Both are jointly trained end to end without separate bridge pretraining. Optimized per-call inference on an H100 achieves p50 latencies of 78\,ms for cognition-only refresh and 25\,ms for action-model invocation. On RoboMME, \pp{} achieves 60.83\% success at the base data scale and 75.54\% with 9$\times$ data, compared with 44.51\% and 57.88\% for FrameSamp+Modul. At base scale, scaling \ponder{} from 0.8B to 9B adds 6.71 percentage points with the \pounce{} architecture unchanged. A separately trained 9B \ponder{} without execution history achieves only 26.21\% under matched supervision. \pp{} also achieves 12.5\% success on RoboCasa-DC and demonstrates real-world applicability on four tasks under asynchronous execution, with 60.98\% mean success versus 40.67\% for FrameSamp+Modul.
\end{abstract}

\section{Introduction}
\label{sec:intro}

Robotic manipulation often depends on evidence absent from the current observation, such as an occluded object, earlier event, referent, or demonstrated procedure~\cite{dai2026robomme,son2026seetraceact}. Pretrained multimodal large language models (MLLMs) can integrate such evidence over long visual histories and reason from examples~\cite{qwen35blog}, yet vision-language-action (VLA) models generally inherit pretrained representations without using this contextual capacity as episode memory. Prior work instead supplies history through purpose-built memory or retrieval~\cite{dai2026robomme,fang2025sam2act,torne2026mem,sridhar2026scaling,shi2026memoryvla}, demonstration encoding~\cite{jain2024vid2robot,kim2025uniskill}, or planning mechanisms~\cite{huang2023voxposer,son2026seetraceact}. We ask whether a pretrained MLLM's native causal context can serve as the episode context engine while an action model handles control.

We evaluate this question on RoboMME, where control requires retaining evidence from execution history or demonstrations, and RoboCasa-DC, where held-out tasks are performed using demonstrations from another embodiment. On RoboMME~\cite{dai2026robomme}, \pizeroonehalf{} reaches 17.93\% without history and 19.73\% with past actions, versus 90.50\% for humans. \pp{} addresses this gap with an asynchronous dual system~\cite{kahneman2011thinking}. \ponder{} (System~2) is a pretrained MLLM that retains episode observations and demonstrations in its causal context and produces a continuous vector called \emph{cognition}. The action model, \pounce{} (System~1), combines the newest ready cognition and its age with current sensory inputs and the instruction. It continues to act while \ponder{} processes new observations. Joint end-to-end training uses no separate bridge pretraining, and the fixed interface lets \ponder{} scale without changing the \pounce{} architecture.

\pp{} outperforms FrameSamp+Modul on RoboMME at both the base and 9$\times$ data scales (Table~\ref{tab:robomme_main}). Removing execution history in a separately trained control with the same 9B \ponder{} and matched supervision lowers base-scale success from 60.83\% to 26.21\%, supporting the contribution of accumulated context beyond model scale and supervision alone (Table~\ref{tab:execution_history}). With full history but no demonstration-reasoning supervision, success remains 48.21\%, exceeding annotation-supervised MemER's 42.38\% (Tables~\ref{tab:robomme_main} and~\ref{tab:cognition_variants}). With full supervision, scaling \ponder{} from 0.8B to 9B improves success by 6.71\,pp while keeping the \pounce{} architecture and interface unchanged (Table~\ref{tab:s2_scale}). \pp{} also achieves 12.5\% success on RoboCasa-DC and 60.98\% mean success across four real-robot tasks under asynchronous execution, supporting applicability across embodiments and beyond simulation (Tables~\ref{tab:robocasa_dc} and~\ref{tab:real}).

\myparagraph{Key contributions.}
\begin{itemize}
    \item \textbf{A pretrained MLLM as an episode context engine.}
          We reuse native causal context to retain execution history and demonstrations, without a purpose-built memory module.

    \item \textbf{A scalable asynchronous context-to-control interface.}
          \ponder{} sends continuous cognition and its age to \pounce{} on independent clocks. The systems are jointly trained end to end without separate bridge pretraining, and their fixed interface allows \ponder{} to scale without changing the \pounce{} architecture.

    \item \textbf{An investigation of context use in robot control.}
          Controlled experiments and ablations on RoboMME examine how execution history, transmitted cognition, model scale, and supervision affect control, while refresh interventions assess dependence on timely cognition. Evaluations on RoboCasa-DC and four real-robot tasks further assess applicability across embodiments and beyond simulation.
\end{itemize}

\section{Related Work}
\label{sec:related}

\myparagraph{Purpose-built episode memory.}
Purpose-built memory systems differ in what state they retain and how it reaches control. FrameSamp+Modul~\cite{dai2026robomme} injects sampled frame tokens through layer-wise modulators. SAM2Act+~\cite{fang2025sam2act} attends to per-view FIFO queues of action-conditioned spatial features. MemER~\cite{sridhar2026scaling} retrieves selected keyframes and emits textual subgoals. MemoryVLA~\cite{shi2026memoryvla} combines per-observation ``cognitive'' and perceptual tokens to retrieve and fuse an external bank. MEM~\cite{torne2026mem} combines short-horizon video with recursive language memory, whereas RoboTTT~\cite{jiang2026robottt} writes history into deployment-time fast weights.

\myparagraph{Demonstrations as test-time context.}
One-shot imitation established that policies can infer tasks from a single example through attention or meta-learned adaptation~\cite{duan2017one,yu2018one}. ViVLA~\cite{chen2025see} jointly models demonstration and robot actions. Vid2Robot~\cite{jain2024vid2robot} and UniSkill~\cite{kim2025uniskill} encode demonstrations through cross-attended prompt-video features and embodiment-agnostic skills, respectively. RoboCat~\cite{bousmalis2023robocat} adapts from action-labeled experience, while SeeTraceAct~\cite{son2026seetraceact} grounds a demonstration-derived latent plan with future visual traces. ICRT~\cite{fu2024context} retains sensorimotor demonstrations and rollouts in a causal transformer's context and predicts actions from the same transformer. \pp{} likewise retains demonstrations and execution history in causal context, but uses a separate pretrained MLLM to produce continuous cognition for the action model.

\myparagraph{Routing context to control.}
Retaining context and routing it to control are distinct choices. Hi Robot~\cite{shi2025hi}, RT-H~\cite{belkhale2024rt}, and Steerable Policies~\cite{chen2026steerable} communicate subgoals, language motions, or coordinates, while LCB~\cite{shentu2024llms} and FiS-VLA~\cite{chen2025fast} transmit internal representations. OpenHelix~\cite{cui2025openhelix} shows that such continuous channels can still collapse toward instruction-level information. Like Helix~\cite{figure2025helix}, \pp{} jointly trains two systems connected by an asynchronous continuous channel. Helix's published System~2 consumes the latest observation, robot state, and language command, whereas \ponder{} retains execution history and demonstrations in persistent MLLM context and passes cognition with its age to \pounce{}. We evaluate this separation through matched history controls, scaling \ponder{} with the \pounce{} architecture fixed, and cognition-refresh interventions.

\section{\pp{}}
\label{sec:method}

\begin{figure*}[!t]
    \vspace*{2mm}
    \centering
    \includegraphics[width=0.8\textwidth]{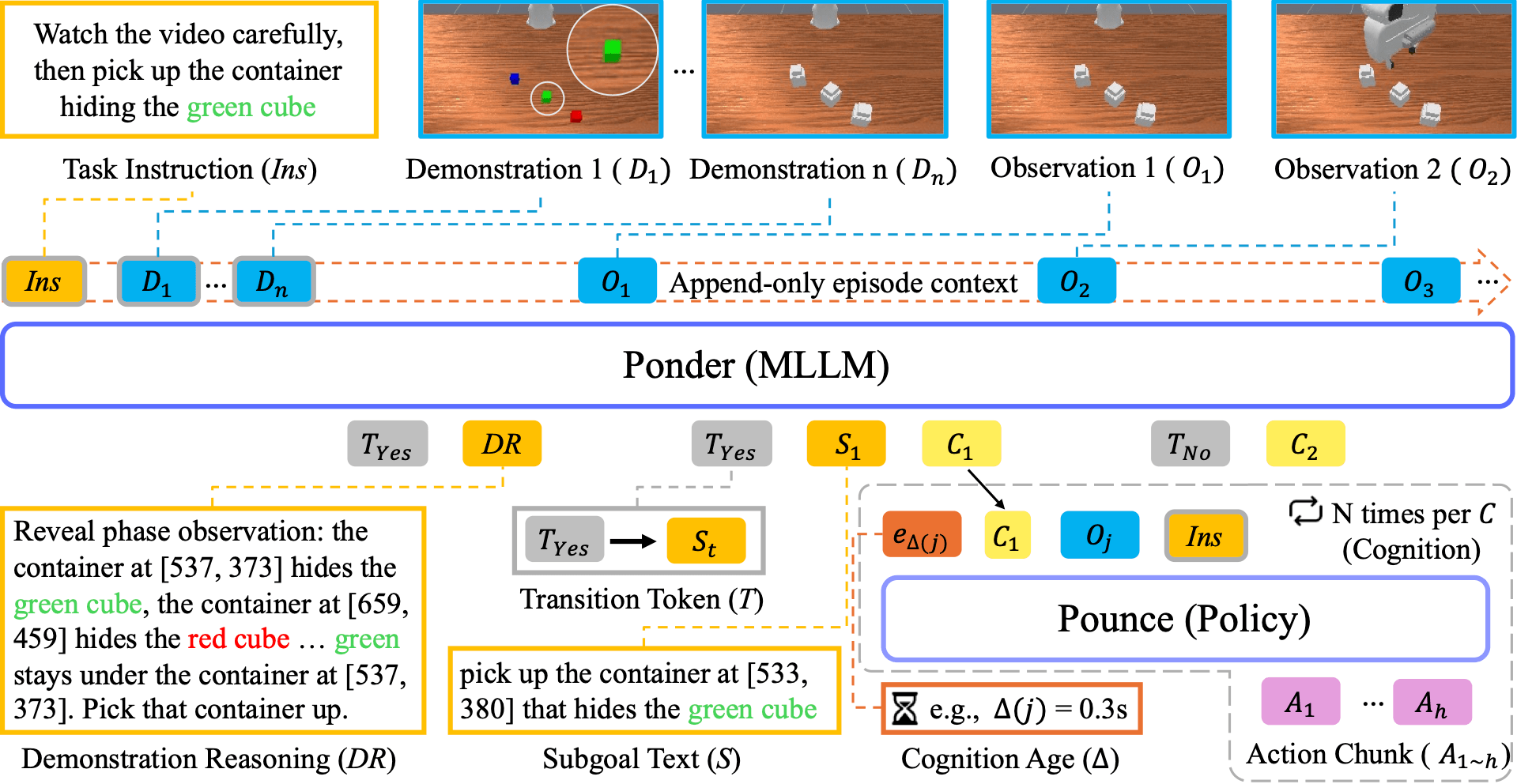}
    \caption{\textbf{\pp{} architecture.} \ponder{} accumulates the instruction $\taskins$, demonstrations $D_{1:n}$, and observations $O_t$. $T_t$ gates optional internal text; carriers form cognition $\mathbf{C}_t$. \pounce{} reuses the newest cognition, combining it with its age and current observation to predict $A_{1:h}$.}
    \label{fig:pp_architecture}
    \vspace{-1em}
\end{figure*}

\pp{} retains episode context in a pretrained MLLM (\ponder{}) and passes a compact representation to an action model (\pounce{}). At each query, \ponder{} appends a new observation, any internally generated text, and $K$ carrier tokens with learned input embeddings. The carriers' final hidden states form the continuous cognition used by \pounce{} to predict action chunks. We train the system end to end using action supervision and, where available, transition, subgoal-text, and demonstration-reasoning targets. Fig.~\ref{fig:pp_architecture} summarizes the architecture and notation.

\subsection{Architecture}
\label{sec:arch}

\myparagraph{Ponder (System~2).} At query $t$, \ponder{} appends observation $O_t$ to an append-only causal context containing the instruction $\taskins$, optional demonstrations $D_{1:n}$, internally generated subgoal text and demonstration reasoning, and earlier cognition carriers. This context lets each query attend to accumulated observations and demonstrations without explicit action history. When trained with grounding supervision, \ponder{} predicts $T_t\in\{\tyes,\tno\}$. If $T_t{=}\tyes$, it generates $\subgoalreason_t$ before inserting the carrier positions. For models trained with demonstration-reasoning targets, $\demoreason$ precedes $\subgoalreason_t$ at the first execution transition of a demonstration episode. If $T_t{=}\tno$, carriers follow $T_t$ without a text payload. Their final hidden states form $\mathbf{C}_t\in\mathbb{R}^{K\times H}$ without decoding or pooling, where $H$ is the MLLM hidden width. Thus every query produces fresh cognition.

\myparagraph{Pounce (System~1).} At invocation $j$, \pounce{} directly receives the instruction $\taskins$, current observation $O_j$, and, where supported, proprioception. A projected prefix supplies the newest ready cognition $\widetilde{\mathbf{C}}_j$ and its age $\Delta(j)$. If no cognition is ready, $\widetilde{\mathbf{C}}_j=\mathbf{C}_{\varnothing}$ for learned null cognition $\mathbf{C}_{\varnothing}\in\mathbb{R}^{K\times H}$ and $\Delta(j)=0$; otherwise, $\widetilde{\mathbf{C}}_j=\mathbf{C}_{\mathrm{ref}(j)}$:
\[
    \mathbf{P}_j = [\,\mathbf{e}_{\Delta(j)},\;\mathbf{W}_c \widetilde{\mathbf{C}}_{j,1},\ldots,\mathbf{W}_c \widetilde{\mathbf{C}}_{j,K}\,].
\]
Here $\mathbf{W}_c$ projects cognition to the action model's width, and $\mathbf{e}_{\Delta(j)}$ is a learned projection of a sinusoidal age encoding. The action head predicts an $h$-action chunk $A_{1:h}$, with only cognition and age crossing the default System~2-to-System~1 interface.

\subsection{Training and Grounding}
\label{sec:training}

We initialize both systems from pretrained checkpoints and jointly optimize all trainable components end to end without separate bridge pretraining. We randomly initialize the carrier input embeddings and cognition and age projectors, and set learned null cognition to zero.
We train on episode-level causal sequences, with gradients propagating through earlier context without temporal detachment.
Grounding denotes optional LM-head supervision of transitions, subgoals, and demonstration reasoning.
The training objective is
\[
    \mathcal{L} = w_1 \mathcal{L}_\text{fm} + w_2 \mathcal{L}_\text{ground},
\]
where $\mathcal{L}_\text{fm}$ is action flow-matching MSE averaged over valid targets and $\mathcal{L}_\text{ground}$ is token cross-entropy over $T_t^*$, $\subgoalreason_t$ at annotated transitions, and, on demonstration episodes with reasoning annotations, first-execution $\demoreason$. The flow-matching loss reaches \ponder{} only through $\mathbf{C}_t$, while grounding enters through the LM head. Multiple \pounce{} invocations can reuse one cognition state and accumulate action gradients there. To limit co-training shortcuts~\cite{ye2026st4vla,cui2025openhelix}, we multiply these gradients into \ponder{} by 0.5 while leaving LM-head gradients unchanged. When no grounding annotations are available, we set $w_2{=}0$.

Let $\mathcal T^*$ denote the annotated transition queries, with targets
\[
    T_t^* =
    \begin{cases}
        \tyes, & t\in\mathcal T^*,\\
        \tno, & t\notin\mathcal T^*.
    \end{cases}
\]
During training, we use teacher forcing for ground-truth transition tokens and their associated text. An ordinary transition uses $[\,\tyes,\subgoalreason_t\,]$, while a non-transition uses $\tno$ without a text payload. For demonstration episodes with reasoning annotations, the first execution transition instead uses $[\,\tyes,\demoreason,\tyes,\subgoalreason_t\,]$, supervising demonstration reasoning once per episode before the first subgoal. When trained with grounding supervision, \ponder{} predicts transitions and generates text at inference. Without LM-head grounding, it bypasses transition and text decoding and produces cognition only. In both cases, the default interface passes only cognition and its age to \pounce{}.

\subsection{Asynchronous Schedule}
\label{sec:schedule}

\ponder{} and \pounce{} run on independent clocks. \pounce{} can continue predicting actions while \ponder{} processes a new observation. It reuses the newest cognition state until a newer \ponder{} query completes. Its age is measured from the source observation. Let $\tau^{(2)}_t$ be the source-observation time of \ponder{} query $t$, $d_t$ its compute delay, and $\tau^{(1)}_j$ the time of \pounce{} invocation $j$. The latest-ready rule selects
\[
    \mathrm{ref}(j)=\max\{t:\tau^{(2)}_t+d_t\leq\tau^{(1)}_j\},
\]
with age $\Delta(j)=\tau^{(1)}_j-\tau^{(2)}_{\mathrm{ref}(j)}$. An empty set selects learned null cognition with age zero.

\subsection{Inference Optimization}
\label{sec:serving}

On a single H100 (bf16, batch 1), append-only \texttt{StaticCache} sessions use a preallocated transformer key-value (KV) cache that retains past keys and values, so each call encodes only newly appended tokens. We set \ponder{}'s context limit to 16K tokens, as no reported evaluation episode reaches it within its budget. Across context lengths from 0.8K to 14K tokens, p95 latency is 93.7\,ms for cognition-only refreshes and 871.1\,ms for forced 45-token subgoal decodes. This native cache is not a separate episode-memory store or retrieval mechanism. Fused Triton kernels~\cite{ma2025running} reduce \pounce{} p50 latency from 142 to 25\,ms ($5.7\times$). Table~\ref{tab:clock_budget} compares eager \pounce{} and per-query context re-encoding with fused kernels and session-resident \texttt{StaticCache}, respectively; the latter also uses \texttt{torch.compile} for cache-only language-model appends. These optimizations meet the 1\,Hz per-call budgets on the profiled hardware. The \pounce{} profile uses a $K{=}8$ cognition prefix, while training and evaluation use $K{=}1$.

\begin{table}[H]
    \centering
    \caption{\textbf{Clock targets and inference latency.} Model rates are inverse p50 latencies, not concurrent throughput. Bold marks optimized rates meeting the target; $\dagger$ marks a rate below target.}
    \label{tab:clock_budget}
    \maintablestyle
    \fittable{%
    \begin{tabular}{llrr}
        \toprule
        Clock & Target & \multicolumn{2}{c}{Rate equivalent (per-call p50)} \\
        \cmidrule(lr){3-4}
              &        & Unoptimized & Optimized \\
        \midrule
        Action playback from chunks & 20\,Hz & \multicolumn{2}{c}{20\,Hz (simulation setting)} \\
        \midrule
        System~1: \pounce{} invocation & 1\,Hz & 7.1\,Hz (142\,ms) & \textbf{40\,Hz} (25\,ms) \\
        \midrule
        System~2: cognition only & 1\,Hz & 1.9\,Hz (521\,ms) & \textbf{12.8\,Hz} (78\,ms) \\
        \hspace{4.6em} +45-token subgoal & 1\,Hz & 0.31\,Hz$^\dagger$ (3.3\,s) & \textbf{1.2\,Hz} (0.82\,s) \\
        \bottomrule
    \end{tabular}%
    }
\end{table}


\begin{table*}[!t]
    \vspace*{2mm}
    \centering
    \caption{\textbf{RoboMME success rate (\%) by memory design.} Each family score is the mean over four tasks. Bold marks the best non-oracle result per scale. $^*$ denotes results reported by RoboMME~\cite{dai2026robomme}. $\dagger$ denotes our 9$\times$-data training run.}
    \label{tab:robomme_main}
    \maintablestyle
    \scalebox{0.92}{%
    \fittable{%
    \begin{tabular}{llrrrrr}
        \toprule
        Method                       & Episode memory                    & Counting       & Permanence     & Reference      & Imitation      & Average        \\
        \midrule
        \multicolumn{7}{l}{\textit{1$\times$ data: no learned episode memory}}                                                                                                                   \\
        \pizeroonehalf$^*$~\cite{intelligence2025pi_}           & None                              & 28.78          & 17.00          & 17.16          & 8.78           & 17.93          \\
        \pizeroonehalf{} + past actions$^*$~\cite{dai2026robomme} & Action history               & 29.09          & 22.75          & 15.92          & 11.17          & 19.73          \\
        \addlinespace
        \multicolumn{7}{l}{\textit{1$\times$ data: purpose-built memory systems}}                                                                                                            \\
        SAM2Act+$^*$~\cite{fang2025sam2act}                 & Per-view FIFO spatial-feature queues & 35.33       & 26.00          & 16.83          & 7.33           & 21.37          \\
        SimpleSG+QwenVL$^*$~\cite{dai2026robomme}          & Subgoal-text list                 & 44.61          & 19.61          & 25.22          & 26.56          & 29.00          \\
        GroundSG+QwenVL$^*$~\cite{dai2026robomme}          & Subgoal-text + bbox lists         & 38.00          & 39.34          & 31.56          & 21.89          & 32.70          \\
        MemER$^*$~\cite{sridhar2026scaling}                    & Selected-keyframe image buffer    & 48.83          & 53.16          & 38.00          & 29.50          & 42.38          \\
        FrameSamp+Modul$^*$~\cite{dai2026robomme}          & Sampled-frame token buffer        & 65.22          & 25.11          & 36.33          & \textbf{51.39} & 44.51          \\
        \addlinespace
        \multicolumn{7}{l}{\textit{1$\times$ data: pretrained MLLM context}}                                                                                                                             \\
        \pp{} (ours)                & Append-only MLLM causal context   & \textbf{74.67} & \textbf{62.83} & \textbf{72.17} & 33.67 & \textbf{60.83} \\
        \midrule
        \multicolumn{7}{l}{\textit{9$\times$ data}}                                                                                                                                                      \\
        FrameSamp+Modul$^\dagger$~\cite{dai2026robomme}    & Sampled-frame token buffer        & \textbf{86.00} & 24.50          & 58.00          & \textbf{63.00} & 57.88          \\
        \pp{} (ours)                & Append-only MLLM causal context   & 81.33 & \textbf{80.17} & \textbf{92.67} & 48.00 & \textbf{75.54} \\
        \midrule
        \multicolumn{7}{l}{\textit{Oracle / human references}}                                                                                                                                           \\
        Human$^*$~\cite{dai2026robomme}                    & Human memory                      & 88.50          & 91.00          & 93.00          & 89.50          & 90.50          \\
        SimpleSG+Oracle$^*$~\cite{dai2026robomme}          & Oracle subgoal                    & 82.56          & 21.56          & 32.28          & 61.94          & 49.58          \\
        GroundSG+Oracle$^*$~\cite{dai2026robomme}          & Grounded oracle subgoal           & 83.86          & 93.31          & 95.16          & 63.98          & 84.08          \\
        \bottomrule
    \end{tabular}%
    }
    }
\end{table*}

\begin{figure*}[!t]
    \centering
    \includegraphics[width=0.8\textwidth]{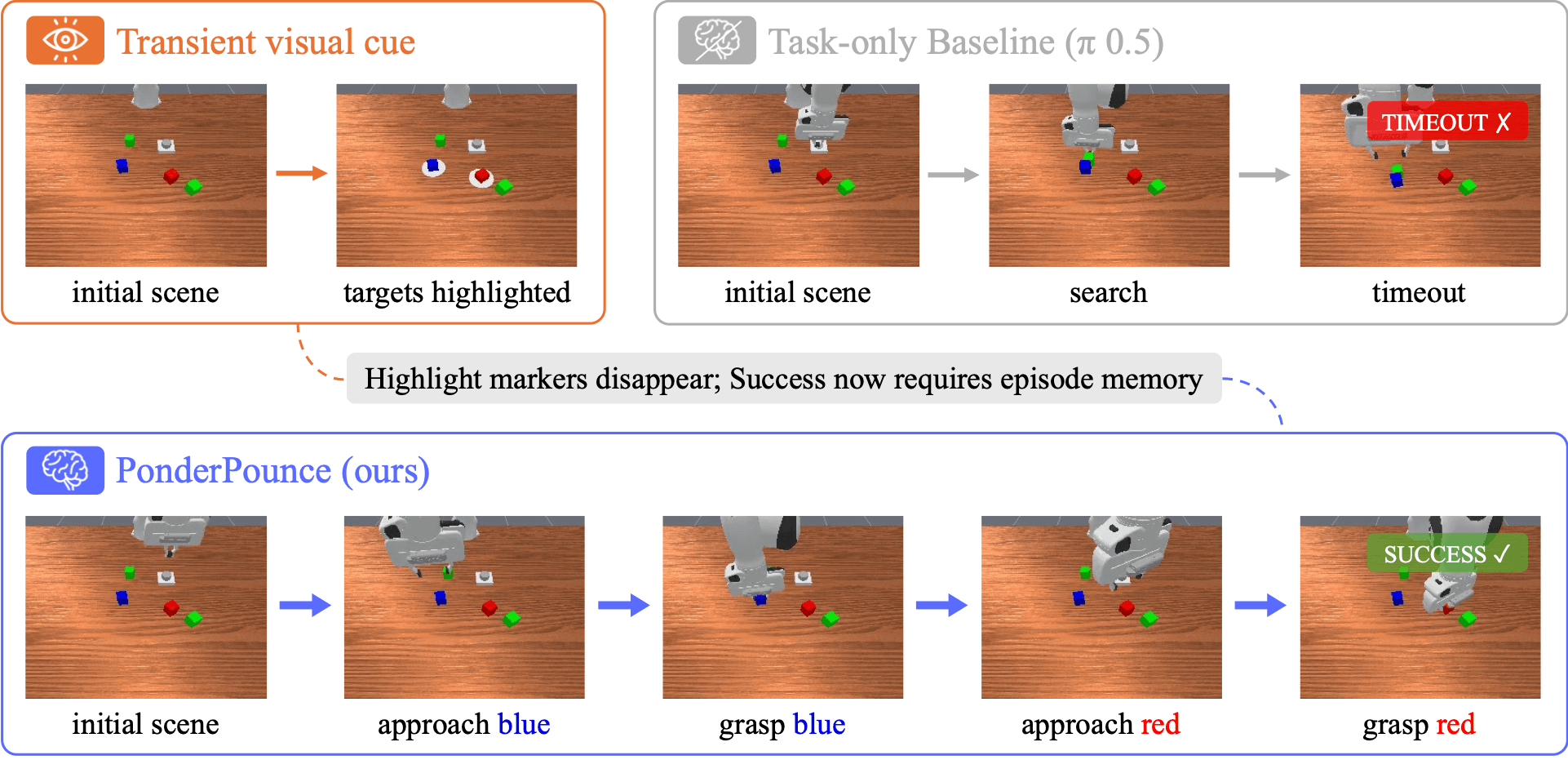}
    \caption{\textbf{Qualitative RoboMME PickHighlight.} The task is \emph{``Press the button, pick all cubes that were highlighted, then stop.''} Pressing the button briefly marks the blue and red targets, after which the markers disappear. \pizeroonehalf searches and times out, whereas \pp{} retains the observed cue, grasps both targets, and presses the button again to complete the task. The final button press is not shown.}
    \label{fig:robomme_qualitative}
    \vspace{-1em}
\end{figure*}

\section{Experiments}
\label{sec:exp_sim}

\subsection{Simulation Setup}
\label{sec:simulation_setup}

\myparagraph{Benchmarks.}
We evaluate memory-dependent control on RoboMME~\cite{dai2026robomme}, whose 16 tasks require information from earlier observations or demonstrations and span four families: Counting, Permanence, Reference, and Imitation. We use 1,587 training episodes at the base scale and a fresh oracle collection of 14,400 episodes at 9$\times$ scale, rather than duplicated data. Grounding annotations are simulator-derived. For episodes with demonstrations, we additionally generate demonstration-reasoning targets from deterministic simulator templates using the same trajectories.

To assess demonstration-conditioned control across embodiments, we use RoboCasa-DC~\cite{son2026seetraceact} in its Category-Balanced / Cross-Embodiment Demonstration setting. The split contains 19 training tasks and five held-out tasks covering pick-and-place, doors, drawers, coffee preparation, and faucet control; each held-out task has a same-family training counterpart. Roughly 100 demonstration--execution pairs per task yield 1,900 training and 500 held-out pairs.

\myparagraph{Model configuration.}
Across both benchmarks, \ponder{} is initialized from Qwen3.5-9B~\cite{qwen35blog}, uses one cognition carrier ($K{=}1$), and is jointly trained with \pounce{}. On RoboMME, \pounce{} is initialized from the 3.6B \pizeroonehalf action model~\cite{intelligence2025pi_}. On RoboCasa-DC, \pounce{} is initialized from the 3B GR00T~N1.5 action model~\cite{bjorck2025gr00t}.

\myparagraph{Training configuration.}
For both benchmarks, we use AdamW with global batch 32, gradient clipping at 1.0, and 100 warmup steps followed by a constant learning rate. Demonstration videos are sampled at 1\,Hz.

On RoboMME, we train for 4,320 episode-based steps with learning rate $2\!\times\!10^{-5}$. The SigLIP vision tower remains frozen, while the PaliGemma language model and action expert are trainable. We use action and grounding loss weights $(w_1,w_2)=(1,0.1)$ without cognition dropout. Action chunks contain 20 steps (1\,s at 20\,Hz). During training, we sample \pounce{} prediction times every 100\,ms on average, with \ponder{} queries averaging 1\,s and simulated compute delays averaging 300\,ms. The base and 9$\times$ runs differ only in training data.

On RoboCasa-DC, we train for 4,000 episode-based steps with learning rate $1\!\times\!10^{-5}$, fully fine-tuning \ponder{} without LoRA. We use action supervision alone, $(w_1,w_2)=(1,0)$. Cognition dropout replaces the transmitted state with learned null cognition on 15\% of training ticks. Action chunks contain 16 steps (0.8\,s at 20\,Hz). Training invokes \pounce{} every 2\,s and \ponder{} every 4\,s with a 300\,ms simulated delay.

\myparagraph{Baseline models.}
On RoboMME, we compare policies without learned episode memory, symbolic policies, and visual-memory methods, alongside human and oracle references (Table~\ref{tab:robomme_main}). The QwenVL baselines and MemER train separate Qwen3-VL-4B predictors on benchmark-provided subgoals. MemER additionally uses keyframe annotations~\cite{dai2026robomme}. On RoboCasa-DC, we compare demonstration-conditioned policies in the cross-embodiment setting, using baseline results reported by SeeTraceAct~\cite{son2026seetraceact} (Table~\ref{tab:robocasa_dc}).

The \pizeroonehalf{} and FrameSamp+Modul baselines use 80K frame-based training steps. The data-matched 9$\times$ FrameSamp+Modul run changes only the training collection.

\myparagraph{Evaluation protocol.}
Each reported \pp{} configuration uses one checkpoint. RoboMME results average three evaluations of 50 episodes per task on the same scenes. RoboMME evaluations synchronize both models at 1\,Hz and start control when cognition is ready, with simulated delay and cognition age fixed at 300\,ms. RoboCasa-DC results average five evaluations of 50 episodes per held-out task. Episodes are capped at 1,300 environment steps on RoboMME and 1,000 on RoboCasa-DC.

\subsection{Simulation Results}

Table~\ref{tab:robomme_main} shows that \pp{} reaches 60.83\% at the base data scale, exceeding FrameSamp+Modul by 16.32\,pp. With 9$\times$ data, success rises by 14.71\,pp to 75.54\%, outperforming the data-matched baseline by 17.66\,pp. At base scale, \pizeroonehalf{} reaches 17.93\%, and action history raises it only to 19.73\%. \pp{} leads on Permanence and Reference at both scales, exceeding FrameSamp+Modul by 55.67 and 34.67\,pp, respectively, at 9$\times$ scale. These are the largest family-level gains, consistent with native context helping hidden-state tracking and reference resolution. FrameSamp+Modul remains stronger on Imitation and 9$\times$ Counting. In PatternLock, we observed path deviations when subgoal updates lagged under 1\,Hz inference, suggesting that coarse temporal resolution contributes to the Imitation gap. Fig.~\ref{fig:robomme_qualitative} illustrates how episode context preserves a transient PickHighlight cue after its visual markers disappear.

Table~\ref{tab:robocasa_dc} reports results on RoboCasa-DC. Trained with action supervision alone, \pp{} reaches 12.5\%, compared with 11.6\% for SeeTraceAct, the strongest published baseline. For the same checkpoint, replacing cognition with the learned null state at inference lowers success to 8.6\%. The drop despite cognition dropout during training suggests reliance on demonstration-conditioned cognition. These results support applicability across action backbones.

\begin{table*}[!t]
    \vspace*{2mm}
    \centering
    \caption{\textbf{Supervision and context-to-control variants.} RoboMME success rate (\%, 1$\times$ data). Continuous-cognition rows are matched supervision ablations. Bold marks the best per column.}
    \label{tab:cognition_variants}
    \maintablestyle
    \scalebox{0.95}{%
    \fittable{%
    \begin{tabular}{lccrrrrr}
        \toprule
        Variant                       & Training & \makecell{LM-head\\targets} & \makecell{Counting} & \makecell{Permanence} & \makecell{Reference} & \makecell{Imitation} & \makecell{Average} \\
        \midrule
        Continuous cognition (ours)   & joint    & \makecell{transition + subgoal text +\\demo reasoning} & \textbf{74.67} & \textbf{62.83} & 72.17          & 33.67          & \textbf{60.83} \\
        \quad w/o demo reasoning      & joint    & transition + subgoal text & 70.50          & 60.00          & 36.67          & 25.67          & 48.21          \\
        \quad w/o LM-head grounding   & joint    & none                  & 49.34          & 27.17          & 21.50          & 13.84          & 27.96          \\
        \midrule
        Subgoal-text reference         & separate & \makecell{transition + subgoal text +\\demo reasoning} & 62.00          & 61.67          & \textbf{73.34} & 42.83          & 59.96          \\
        FrameSamp+Modul~\cite{dai2026robomme} & --       & --                    & 65.22          & 25.11          & 36.33          & \textbf{51.39} & 44.51          \\
        \bottomrule
    \end{tabular}%
    }
    }
    \vspace{-1em}
\end{table*}

\begin{table}[!t]
    \centering
    \caption{\textbf{RoboCasa-DC success rate (\%).} \pp{} reports mean $\pm$ s.d. over five evaluations of one checkpoint. Baseline scores are from SeeTraceAct~\cite{son2026seetraceact}.}
    \label{tab:robocasa_dc}
    \maintablestyle
    \scalebox{0.95}{%
    \fittable{%
    \begin{tabular}{lr}
        \toprule
        Method                    & Success rate \\
        \midrule
        Vid2Robot~\cite{jain2024vid2robot} & 8.8                  \\
        UniSkill~\cite{kim2025uniskill} & 11.2                 \\
        ViVLA~\cite{chen2025see}    & 8.0                  \\
        SeeTraceAct~\cite{son2026seetraceact} & 11.6                 \\
        \midrule
        \pp{} (ours)              & 12.5 $\pm$ 0.9       \\
        \pp{}, cognition disabled & 8.6 $\pm$ 0.4        \\
        \bottomrule
    \end{tabular}
    }
    }
    \vspace{-1.5em}
\end{table}

\subsection{Investigating Context Use in Robot Control}
\label{sec:analysis}

Latent channels can collapse to instruction-level summaries that are insensitive to visual changes~\cite{cui2025openhelix}. This motivates examining whether execution history contributes beyond model scale and grounding supervision, how supervision affects control, and how continuous cognition compares with decoded subgoal text in control performance. Table~\ref{tab:clock_budget} compares per-call costs of cognition refresh and text decoding. We also test whether pretrained System~2 capacity transfers through a fixed context-to-control interface and examine cognition refresh and staleness in Sec.~\ref{sec:staleness}.

\myparagraph{Execution history under matched supervision.}
\label{sec:execution_history}
We separately train a current-observation-only control with the same 9B \ponder{}, \pizeroonehalf{} \pounce{}, 1$\times$ dataset, and transition/subgoal/demo-reasoning supervision, matching the loss weights, learning rate, global batch, update count, and asynchronous schedule. During both training and evaluation, each query retains the task, demonstrations, and current observation but excludes earlier execution observations, generated text, and cognition. Episode-level sampling, action targets, and current-query text generation remain unchanged.

Success falls from 60.83\% with full history to 26.21\% without it (Table~\ref{tab:execution_history}), supporting a contribution from accumulated execution context beyond model scale and grounding supervision. This comparison does not isolate the individual contributions of past images, generated text, and cognition.

\myparagraph{Supervision and interface form.}
\label{sec:reasoning_supervision}
To assess the role of grounding supervision in control, we first remove demonstration-reasoning targets while retaining transition and subgoal supervision, reducing success from 60.83\% to 48.21\% (Table~\ref{tab:cognition_variants}). Even without demonstration reasoning, this variant exceeds the annotation-supervised SimpleSG+QwenVL (29.00\%), GroundSG+QwenVL (32.70\%), and MemER (42.38\%), as well as FrameSamp+Modul (44.51\%), the strongest non-oracle baseline, showing that the gains are not attributable to demonstration reasoning alone. However, removing all LM-head grounding reduces success to 27.96\%, below all four baselines, suggesting that grounding helps exploit pretrained context for control.

\begin{table}[!t]
    \centering
    \caption{\textbf{Execution-history control on RoboMME (\%, 1$\times$ data).} Matched task/demo context and grounding supervision.}
    \label{tab:execution_history}
    \maintablestyle
    \fittable{%
    \begin{tabular}{lrrrrr}
        \toprule
        Execution context & Counting & Permanence & Reference & Imitation & Average \\
        \midrule
        Full history & 74.67 & 62.83 & 72.17 & 33.67 & 60.83 \\
        Current obs. only & 26.83 & 23.33 & 33.17 & 21.50 & 26.21 \\
        \bottomrule
    \end{tabular}%
    }
\end{table}

For the subgoal-text reference, we use a separately trained \ponder{} and pass its generated subgoal text, rather than cognition, to a subgoal-conditioned \pizeroonehalf{}. The subgoal prompt updates only at predicted transitions and persists between them, while demonstration reasoning remains internal to \ponder{}. This reference reaches 59.96\% versus 60.83\% for continuous cognition. Despite similar success rates, the continuous interface propagates action-loss gradients into \ponder{} during end-to-end training and makes updated context available to \pounce{} after every \ponder{} query, including within a subgoal.

\myparagraph{Does pretrained context capacity transfer to control?}
We replace Qwen3.5-9B with Qwen3.5-0.8B and adapt the cognition projection to its hidden width, keeping the \pounce{} architecture and projected interface unchanged. The pretrained 9B model reaches 60.83\%, 6.71\,pp above the 0.8B model at 54.12\% (Table~\ref{tab:s2_scale}), showing that a larger context engine can improve control while off the fast action path. The randomly initialized 9B model scores 0.00\% with the same recipe, indicating recipe-specific reliance on pretraining rather than a matched-optimization comparison.

\subsection{Cognition Refresh and Staleness}
\label{sec:staleness}
We examine whether cognition must be refreshed within a subgoal and how sensitivity to stale cognition changes with the training refresh interval.

\begin{table}[!t]
    \centering
    \caption{\textbf{RoboMME success rate by System~2 scale and initialization (\%, 1$\times$ data).} The \pounce{} architecture and interface are unchanged across model sizes.}
    \label{tab:s2_scale}
    \maintablestyle
    \scalebox{0.95}{%
    \fittable{%
    \begin{tabular}{lcrrrrr}
        \toprule
        \ponder{} (System~2) & Init. & Counting & Permanence & Reference & Imitation & Average \\
        \midrule
        Qwen3.5 9B          & pretrained     & 74.67 & 62.83 & 72.17 & 33.67 & \textbf{60.83} \\
        Qwen3.5 0.8B        & pretrained     & 67.00 & 58.00 & 64.50 & 27.00 & 54.12 \\
        Qwen3.5 9B          & random         & 0.00 & 0.00 & 0.00 & 0.00 & 0.00 \\
        \bottomrule
    \end{tabular}
    }
    }
    \vspace{-1em}
\end{table}

\myparagraph{Within-subgoal cognition refresh.}
To probe the role of cognition beyond a subgoal summary, we test whether the full-history 9B checkpoint tolerates cognition delivery only at predicted transitions. In this test, \ponder{} still runs at every query, but \pounce{} receives new cognition only at the first query and predicted subgoal transitions. Between transitions, it reuses the last state with either its true age or an age fixed at 300\,ms to present held cognition as fresh, matching the age input in standard RoboMME inference. With transition-only delivery, success falls from 60.83\% to 1.83\% when age is fixed at 300\,ms, but recovers to 22.42\% with true age (Table~\ref{tab:cognition_refresh}). These results are consistent with \pounce{} treating outdated execution context as current when held cognition is presented as fresh. Reporting its true age mitigates this mismatch but does not supply updated context, leaving performance below regular refresh.

\begin{table}[!t]
    \vspace*{2mm}
    \centering
    \caption{\textbf{Within-subgoal cognition refresh.} RoboMME success rate (\%, 1$\times$ data) using the same checkpoint.}
    \label{tab:cognition_refresh}
    \maintablestyle
    \scalebox{0.95}{%
    \fittable{%
    \begin{tabular}{llrrrrr}
        \toprule
        Refresh & Cognition / age & Counting & Permanence & Reference & Imitation & Average \\
        \midrule
        Every query & latest $\mathbf{C}_t$ / 300\,ms & 74.67 & 62.83 & 72.17 & 33.67 & 60.83 \\
        Transitions only & held $\mathbf{C}$ / true age & 19.83 & 15.50 & 33.33 & 21.00 & 22.42 \\
        Transitions only & held $\mathbf{C}$ / 300\,ms & 0.17 & 0.17 & 0.67 & 6.33 & 1.83 \\
        \bottomrule
    \end{tabular}%
    }
    }
    \vspace{-1em}
\end{table}

\begin{figure}[!t]
    \centering
    \includegraphics[width=0.85\linewidth]{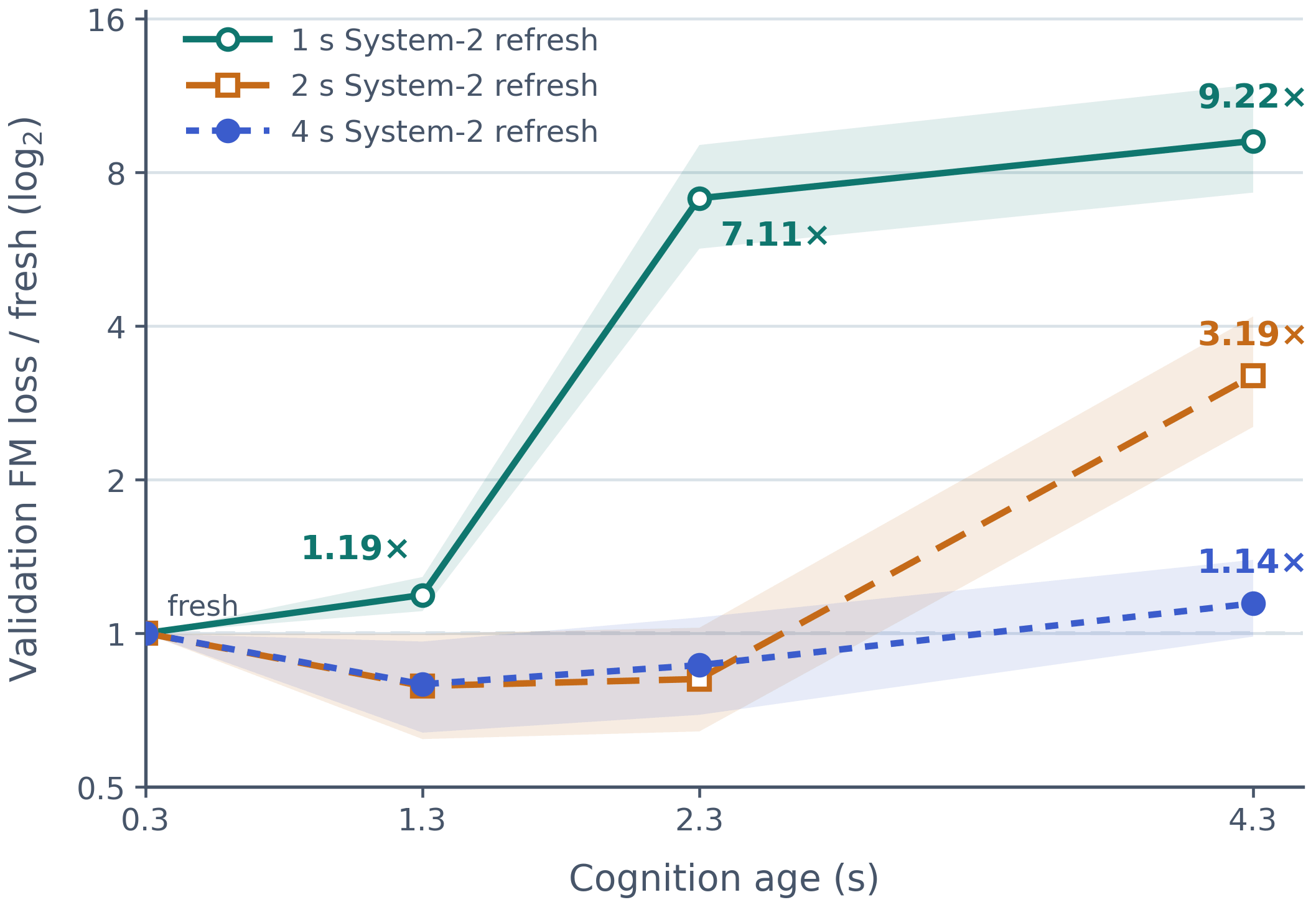}
    \caption{\textbf{Measured staleness under slow-refresh training.} The curves are checkpoints trained with 1, 2, or 4\,s System~2 refresh intervals. Bands show 95\% task-stratified bootstrap CIs. The log$_2$ axis shows validation flow-matching loss normalized by each checkpoint's own 0.3\,s condition. Lower is better. This measures offline loss, not closed-loop success.}
    \label{fig:cognition_staleness}
    \vspace{-1em}
\end{figure}

\myparagraph{Sensitivity to stale cognition.}
We evaluate stale-cognition sensitivity offline on the same 757 held-out ticks from 80 task-balanced RoboMME episodes using teacher forcing (Fig.~\ref{fig:cognition_staleness}). We replace $\mathbf{C}_t$ with $\mathbf{C}_{t-k}$ for $k\in\{0,1,2,4\}$ and report its true age of $0.3+k$ seconds, keeping all other inputs, targets, flow noise, and diffusion times fixed.
For the 1\,s-refresh checkpoint, normalized loss rises from $1.00\times$ at 0.3\,s to $7.11\times$ at 2.3\,s and $9.22\times$ at 4.3\,s. Training with 2 and 4\,s refresh intervals reduces the 4.3\,s ratios to $3.19\times$ and $1.14\times$, respectively, but increases fresh-cognition (0.3\,s) absolute loss from 0.117 to 0.213 and 0.232. These offline losses do not directly predict closed-loop success rates.

\begin{table}[t]
    \vspace*{2mm}
    \centering
    \caption{\textbf{Real-robot asynchronous deployment timing.} Percentiles pool timing samples across all 59 trials.}
    \label{tab:real_timing}
    \maintablestyle
    \scalebox{0.90}{%
    \fittable{%
    \begin{tabular}{llrr}
        \toprule
        System & Metric (ms) & Median & p95 \\
        \midrule
        System~1 (\pounce{}) & Invocation interval & 1,000 & 1,001 \\
        & Compute time & 181 & 241 \\
        & Non-null cognition age & 1,803 & 2,006 \\
        \midrule
        System~2 (\ponder{}) & Invocation interval & 1,000 & 1,001 \\
        & Compute time & 187 & 537 \\
        \midrule
        Action delivery & Observation age & 776 & 1,223 \\
        \bottomrule
    \end{tabular}%
    }
    }
    \vspace{-1.2em}
\end{table}

\begin{table*}[!t]
    \vspace*{2mm}
    \centering
    \begin{minipage}[c]{0.68\textwidth}
        \centering
        \caption{\textbf{Real-robot success rate (\%).} Entries show successes/trials (rounded \%).\\Average is the unweighted mean of unrounded task rates.}
        \label{tab:real}
        \maintablestyle
        \resizebox{0.98\linewidth}{!}{%
        \begin{tabular}{lccccc}
            \toprule
            Method & \makecell{Put Fruits\\(Counting)} & \makecell{Track Cube\\(Permanence)} & \makecell{Repick Block\\(Reference)} & \makecell{Draw Pattern\\(Imitation)} & Average \\
            \midrule
            \pizeroonehalf{}   & 9/14 (64\%) & 4/16 (25\%) & 0/14 (0\%) & 1/15 (7\%) & 23.99\% \\
            FrameSamp+Modul    & 7/14 (50\%) & 5/16 (31\%) & 3/14 (21\%) & \textbf{9/15 (60\%)} & 40.67\% \\
            \pp{} (ours)       & \textbf{11/14 (79\%)} & \textbf{10/16 (63\%)} & \textbf{6/14 (43\%)} & \textbf{9/15 (60\%)} & \textbf{60.98\%} \\
            \bottomrule
        \end{tabular}%
        }
    \end{minipage}%
    \begin{minipage}[c]{0.3\textwidth}
        \centering
        \includegraphics[width=0.7\linewidth]{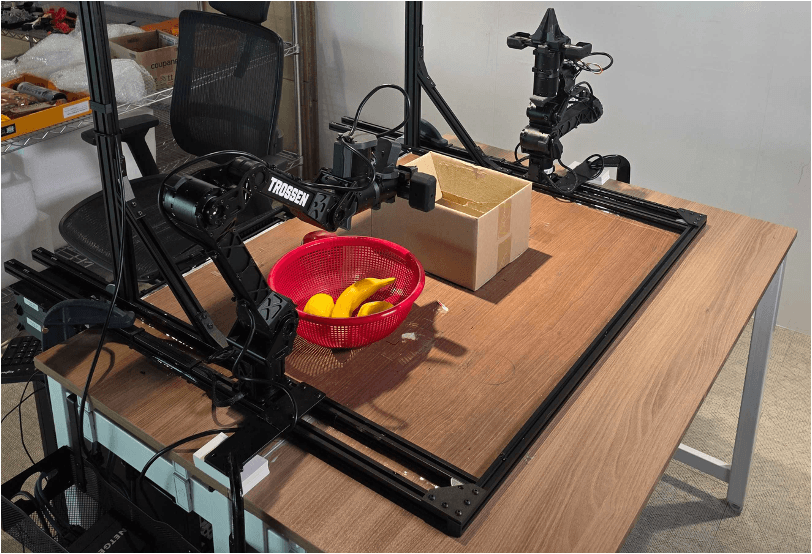}
        \makeatletter
        \def\@captype{figure}
        \makeatother
        \caption{Trossen ALOHA Stationary AI Kit performing PutFruits, using only the right arm.}
        \label{fig:real_platform}
    \end{minipage}
    \vspace{-1em}
\end{table*}

\subsection{Real-Robot Experiments}
\label{sec:exp_real}

\myparagraph{Platform and tasks.}
We adapt the four real-world task designs of RoboMME~\cite{dai2026robomme} to the Trossen ALOHA Stationary AI Kit, using only its right arm for manipulation (Fig.~\ref{fig:real_platform}). Policies use six joint positions and one gripper value for state and action, with external and wrist RGB views. Joint-target action playback is configured for 30\,Hz, separately from model inference. \textit{PutFruits} (Counting) requires transferring an instruction-specified number of toy fruits from a basket into a bin and pressing a stop button. \textit{TrackCube} (Permanence) requires selecting the cup hiding the cube of the specified color based on an earlier video. \textit{RepickBlock} (Reference) requires picking the blocks in the order shown in the demonstration video, and \textit{DrawPattern} (Imitation) requires reproducing a demonstrated path. TrackCube, RepickBlock, and DrawPattern receive a pre-execution video prefix, whereas PutFruits relies only on execution history.

\myparagraph{Data and models.}
All three methods are trained on the same 710 teleoperated episodes and evaluated on 59 held-out scenes. Training-episode/evaluation-scene counts are 104/14 for PutFruits, 195/16 for TrackCube, 206/14 for RepickBlock, and 205/15 for DrawPattern. The three demonstration-conditioned tasks use bundled clips, with both evaluation scenes and clips excluded from training. The training data include human-annotated subgoals for \ponder{} supervision but no demonstration-reasoning annotations. \pp{}, FrameSamp+Modul, and \pizeroonehalf{} follow their respective RoboMME training configurations in Sec.~\ref{sec:simulation_setup}. All three methods use 30-step action chunks.

\myparagraph{Protocol.}
\ponder{} and \pounce{} target 1\,Hz on separate H100 GPUs under wall-clock asynchronous execution. \ponder{} uses session-resident \texttt{StaticCache} without \texttt{torch.compile}, and \pounce{} runs eagerly without fused kernels. \pounce{} uses the latest ready cognition and age, with decaying-velocity extrapolation after chunk exhaustion. Table~\ref{tab:real_timing} includes transition-triggered text decoding in \ponder{} timings. Ages span source-observation arrival to \pounce{} invocation or server action emission, not actuation. Per-task client tick rates average 22.9--27.8\,Hz over full episodes. Trials have a 2\,min limit, with a human judging success or failure.

\myparagraph{Results.}
Table~\ref{tab:real} shows that \pp{} achieves a four-task mean success rate of 60.98\%, compared with 40.67\% for FrameSamp+Modul and 23.99\% for \pizeroonehalf{}. \pp{} leads on PutFruits, TrackCube, and RepickBlock, tying FrameSamp+Modul at 9/15 on DrawPattern. The largest gain over FrameSamp+Modul is on TrackCube, from 5/16 to 10/16 successes, consistent with strong Permanence results in simulation. This demonstrates real-world applicability under wall-clock asynchronous execution.

\section{Conclusion}
\label{sec:conclusion}

\pp{} reuses a pretrained MLLM's native causal context as episode memory, asynchronously routing continuous cognition to an action model without a purpose-built memory module. It achieves 60.83\% success on RoboMME at the base data scale and 75.54\% with 9$\times$ data. Removing execution history under matched backbones and supervision reduces success to 26.21\%, supporting the contribution of accumulated context. Scaling \ponder{} improves control without changing the \pounce{} architecture or interface. On RoboCasa-DC, 12.5\% success provides initial evidence of applicability across action-model backbones, while 60.98\% mean success across four real-robot tasks under wall-clock asynchronous execution demonstrates applicability beyond simulation. Together, these results support using pretrained MLLM context for memory-dependent robot control. The context engine adds training and inference costs, and closed-loop robustness to additional delays remains untested.

\section{Future Work}
\label{sec:future_work}

Future work should compare memory architectures under matched backbones, supervision, and training budgets, using history-component ablations and counterfactual episodes to isolate the contribution of execution history. Representation probes could further clarify what cognition encodes and how \pounce{} uses it. For deployment, controlled latency and jitter experiments could characterize robustness beyond the measured conditions. Adaptive cognition delivery and smaller or distilled context models could reduce computation while preserving control performance. To scale training, vision-language-generated or propagated labels may reduce annotation costs, subject to validation of label quality~\cite{xiao2022robotic,zhao2025training,feng2026procvlm}. Scaling soft-token learning may also require longer, temporally coherent vision--action data in which behavior depends on memory and long-horizon goals, with desktop and game interaction providing useful complementary settings~\cite{choi2026d2e,magne2026nitrogen}.

\section*{Acknowledgments}
This work was supported by Institute of Information \& communications Technology Planning \& Evaluation (IITP) grant funded by the Korea government (MSIT) (No. RS-2026-25522885, Development of a World Foundation Model for Training and Deployment of Physical AI Systems).

\bibliographystyle{IEEEtran}
\bibliography{references}

\end{document}